\documentclass[acmlarge,screen,nonacm]{acmart}

\usepackage{array}
\usepackage{float}
\newcolumntype{A}{>{\centering\arraybackslash}m{2cm}}
\newcolumntype{Q}{>{\centering\arraybackslash}m{2cm}}
\newcolumntype{W}{>{\centering\arraybackslash}m{3cm}}
\newcolumntype{E}{>{\centering\arraybackslash}m{2cm}}

\usepackage{etoolbox}
\usepackage{multirow}
\usepackage{url}
\preto\tabular{\setcounter{magicrownumbers}{0}}
\newcounter{magicrownumbers}
\newcommand\rownumber{\stepcounter{magicrownumbers}\arabic{magicrownumbers}}

\setcopyright{none}
\renewcommand\footnotetextcopyrightpermission[1]{}
\makeatletter
\let\@authorsaddresses\@empty
\makeatother

\begin{document}
	
	\title{Part-of-Speech Tagging of Odia Language Using statistical and Deep Learning-Based Approaches}
	
	%\email{webmaster@marysville-ohio.com}
	%\affiliation{%
		%  \institution{Institute for Clarity in Documentation}
		%  \streetaddress{P.O. Box 1212}
		%  \country{USA}
	%\author{Lars Th{\o}rv{\"a}ld}
	%\affiliation{%
		%  \institution{The Th{\o}rv{\"a}ld Group}
		%  \streetaddress{1 Th{\o}rv{\"a}ld Circle}
	%\email{larst@affiliation.org}
	%\affiliation{%
		%  \institution{Inria Paris-Rocquencourt}

	
	\begin{abstract}
		Automatic Part-of-speech (POS) tagging is a preprocessing step of many natural language processing (NLP) tasks such as name entity recognition (NER), speech processing, information extraction, word sense disambiguation, and machine translation. It is a process of assigning the grammatical class label to each word in a sentence with their respective part-of-speech based on morphological and contextual language information. It has already gained a promising result in English and European languages, but in Indian languages, particularly in Odia language, it is not yet well explored because of the lack of supporting tools, resources, and morphological richness of language. Unfortunately, we were unable to locate an open source POS tagger for Odia, and only a handful of attempts have been made to develop POS taggers for Odia language. Odia is one of the official languages of India that comes under Indo Aryan language family, spoken in Odisha and parts of West Bengal, Jharkhand and Chhattisgarh, with around 45 million native speakers. The main contribution of this research work is to present a conditional random field (CRF) and deep learning-based approaches (CNN and Bidirectional Long Short-Term Memory) to develop Odia part-of-speech tagger. We used a publicly accessible corpus as part of the Indian Languages Corpora Initiative (ILCI) phase-II project, which was initiated by the Ministry of Electronics and Information Technology (MeitY) Government of India. The dataset is annotated with the Bureau of Indian Standards (BIS) tagset. However, most of the languages around the globe have used the dataset annotated with Universal Dependencies (UD) tagset. Hence, to maintain uniformity Odia dataset should use the same tagset. So we have constructed a simple mapping from BIS tagset to UD tagset. CRF, Bi-LSTM, and CNN models are trained using ILCI corpus with BIS and UD tagset. We experimented with various feature set inputs to the CRF model, observed the impact of constructed feature set. The deep learning-based model includes Bi-LSTM network, CNN network, CRF layer, character sequence information, and pre-trained word vector. Character sequence information was extracted by using convolutional neural network (CNN) and Bi-LSTM network. Six different combinations of neural sequence labelling models are implemented, and their performance measures are investigated. It has been observed that Bi-LSTM model with character sequence feature and pre-trained word vector achieved a significant state-of-the-art result.
		
		% Unfortunately, we did not find any openly available Odia POS tagger, and only a few attempts have been made to develop POS taggers for Odia.
		
	\end{abstract}

 \author{Tusarkanta Dalai}
\affiliation{%
	\institution{Department of Computer Science and Engineering, NIT Rourkela}
\city{Rourkela}
\country{India}}

%\email{tusarkantadalai@gmail.com}
\author{Tapas Kumar Mishra}
\affiliation{%
	\institution{Department of Computer Science and Engineering, NIT Rourkela}
	\city{Rourkela}
	\country{India}}
%%\affiliation{Department of Computer Science and Engineering, NIT Rourkela}
\author{Pankaj K Sa}
\affiliation{%
	\institution{Department of Computer Science and Engineering, NIT Rourkela}
	\city{Rourkela}
	\country{India}}
%%\affiliation{Department of Computer Science and Engineering, NIT Rourkela}
%%\email{a@g}
	\maketitle

	\section{Introduction}
	\label{intro}

POS tagging is one of the sequence labelling tasks that involves assigning a grammatical category label to each word based on linguistic and contextual information \cite{brill1995transformation}. The tag or label of a word provides information about the word and its surrounding lexical categories. In general, a POS tagger will classify the sentence into several subcategories based on the its parts of speech including noun, pronoun, adjective, verb, adverb and so on. POS tags are useful because they provide linguistic information about how words can be employed in a phrase, sentence, or document. In the field of language processing, POS tagging is an essential pre-processing step for many natural language processing (NLP) frameworks. These NLP frameworks include speech recognition, sentiment analysis, named entity recognition, question answering, word sense disambiguation, and chunking \cite{mitkov2022oxford}. Determining the grammatical class of a language is challenging because it is difficult to assign tags to each word of a sentence when some words have more than one grammatical POS label. In addition syntax or semantics of many languages may differ, and the grammatical structure of many languages varies depending on the context in which it is used. In English, European languages, and other South Asian languages, the problem of POS tagging has been thoroughly investigated. However, there is still a need for research on Indian languages, particularly the Odia language, as it is difficult to do research on languages with complicated morphological inflection and flexible word order. In addition, the lack of capitalization, gender information, and other aspects complicates POS tagging in Odia.
	
	India is the birthplace of five diverse language groups, including Dravidian, Indo-Aryan, Tibeto-Burman, Austro-Asian, and Andamanese. Odia, which belongs to the Indo-Aryan language family and is also known as Oriya, is spoken in India. It is the state language of Odisha, an Indian state. Odia is one of India's 22 official languages and 14 regional languages, and it is the sixth Indian language to be recognized as a classical language. Previously, it was a Scheduled Language under the Eighth Schedule of the Indian Constitution. Odia language originated in the tenth century and later the language was changed under the influence of the Sanskrit language in the sixteenth and seventeenth centuries. Like most of the Indian languages, Odia is also a resource-poor language with less computerization. Some of the NLP tools that have been developed are either not available online or have not been made public or half-furnished. Considering the NLP situation in Indian languages, they are poor so far as the availability of electronically annotated written corpora is concerned. In the field of computational linguistics, there has been a lot of research on other Indian languages, but not much on the Odia language. So, the Odia language needs to be explored from the language processing prospective. Therefore, we researched into Odia POS labelling task from an NLP perspective. Some research articles on NLP have recently been introduced in Odia, which will be discussed in the following units below.

	Many steps are involved in POS tagging, including designing a tagset, creating a corpus by tagging the text of a given language, constructing an algorithm to classify the given un-tagged text, and so on. A corpus is an extensive collection of texts or words that is required for any NLP task, and it can be divided and used for training, validation and testing. It is also difficult to obtain standard corpora in many languages; therefore, datasets or corpora are often a challenge. So, many researchers have constructed their own corpus. For the purpose of our experiment, we have collected publicly available corpus through the website of the Technology development for Indian languages     (TDIL). This was done as part of the project Indian Languages Corpora Initiative phase-II, which was launched by the Ministry of Electronics and Information Technology of the Government of India at Jawaharlal Nehru University in New Delhi. The corpus is tagged using BIS tagset contains 38 tags. The tagsets consist of a collection of tags representing the grammatical class information of a specific language. Tags that describe grammatical categories or POS, such as nouns, verbs, adjectives, pronouns, etc. Several POS tagsets have been used for different languages by a number of researchers or research groups. Various tagset schemes are developed for Indian Languages \cite{chandra2014various} . Bureau of Indian Standards (BIS) tagset, Indian Language Machine Translation (ILMT) tagset, Linguistic Data Consortium for Indian Languages (LDC-IL) tagset, JNU-Sanskrit tagset (JPOS), Sanskrit consortium tagset (CPOS), and Tamil tagset are some tagsets used for Indian languages. However, since all other datasets used in the current experiments all around the globe have been annotated with Universal Dependencies tagset, it was necessary that Odia dataset also uses the same tagset to maintain uniformity. Therefore we have also used UD tagset for tagging the corpus by using a simple mapping from BIS tagset. 
	
	A POS tagging technique is necessary to identify unique grammatical POS for each input word. Several algorithms are used for the POS tagging task, including rule-based, probabilistic, deep learning, and hybrid approaches. Therefore, the system should be able to automatically produce the output as tagged terms from the given input text using the technique. We employed supervised learning method based on conditional random fields (CRF), deep learning approaches like CNN and bidirectional long short-term memory (Bi-LSTM), and a combination of deep learning methods with CRF.
	
	The following are the main contributions of this article:
	\begin{enumerate}
		\item Design the Odia POS corpus using UD tagset.
		\item Develop Odia POS tagger using CRF.
		\item Generate word embedding vector through fastText word embedding method for Odia POS tagging task.
		\item Develop the first deep learning-based Odia POS tagger using CNN and Bi-LSTM.
	\end{enumerate}
	
	The remaining parts of this article are organized in the following manner: Section \ref{related_work} describes literature survey on POS tagging; Section \ref{corpus} presents the Odia POS dataset and UD tagset for Odia language; Section \ref{model} describes the methodologies used to develop POS tagging in the Odia language; Section \ref{result} presents the experimental results of different models; Section \ref{analysis} describes the performance analysis; Section \ref{conclusion} presents the conclusions and discuss future works in these dimensions.
	
	%POS tagging is a key component of natural language processing (NLP).
	
	\section{Related Work}
	
\label{related_work}
	This section presents the existing related work of POS tagging on different languages and presently available work on different approaches. Harris \cite{harris1962string} was the first to explore POS tagging, and he used a rule-based approach to develop POS tagger. Later, numerous rule-based systems were developed to improve the accuracy and efficiency of POS taggers. However, rule-based systems have significant drawbacks, including a high level of human effort, time consumption, and less learning potential. In the nineties, statistical-based machine learning methods were quite popular. Cutting et al. \cite{cutting1992} have proposed a statistical-based Hidden Markov Model (HMM) to develop a POS tagger using Brown Corpus with 50,000 tagged words. In 2001 Lafferty et al. \cite{lafferty2001conditional} introduced another statistical model for POS tagging called conditional random field (CRF), and they observed that CRF performs better than other statistical models in POS tagging tasks. Later neural network-based POS tagging was introduced. It includes CNN, LSTM, Bi-LSTM, RNN, GRU, etc., and these methods are recently used for sequence labelling tasks. In 1994, Helmut \cite{schmid1994part} introduced a neural network-based POS tagging.
	
	Here we discuss some of the recent research work reported on POS tagging in different languages. In paper \cite{dos2014learning} the authors introduced a deep neural-based POS tagger, and the model combined the word and character-level representations. This model achieved good accuracy in the English and Portuguese languages. A Bi-LSTM model was proposed by W. Ling et al.\cite{ling2015finding}. Here Bi-LSTM network was used to compose the characters to construct character representation of words, and it achieved good accuracy in morphologically rich languages. In \cite{plank2016multilingual} the authors developed a multilingual POS tagging with Bi-LSTM. The model was experimented on 22 different languages and achieved a state-of-art performance. This model was performed well by using word embedding and character embedding features. Huang et al. 
	\cite{huang2015bidirectional}  proposed a Bi-LSTM-CRF model for sequence labelling tasks. Here LSTM network is composed with a CRF layer and improves the model's performance. This model has produced a state-of-art result on POS, NER, and chunking datasets.
	Shrivastava and Bhattacharyya \cite{shrivastava2008hindi} proposed  HMM-based tagger. They achieved an accuracy of 93.12\%. And they used a corpus of size 81,751 tokens for the experiment. Alam et al. \cite{alam2016bidirectional} developed a POS tagger for Bangla language using LSTM network and CRF layer with word embedding feature. This model was implemented using publicly available corpus achieved an accuracy 86.0\%. In \cite{krishnan2017character} the author presented a POS tagger for Tamil using character-based Bi-LSTM. They got an accuracy of 86.45\%. Priyadarshi and Saha \cite{priyadarshi2020towards} proposed a Maithili POS tagger using CRF and word embedding. They have used an annotated corpus of 52,190 words and achieved an accuracy of 85.88\%. Warjri et al. \cite{warjri2021part} proposed a Khasi POS tagger using deep learning-based approach on designed Khasi corpus. They achieved an accuracy of 96.98\% by using Bi-lSTM with CRF layer.

	A few efforts have been proposed for the Odia language to build POS tagger. \cite{das2014novel} Das and Patnaik  presented an implementation of a single neural network-based POS tagging algorithm trained on a manually-annotated corpus. A small tag set of five tags was considered for labelling the data. Voting-based selection rule to obtain the output and forward propagation method was used for error correction. The Odia POS tagger using artificial neural network (ANN) obtained an accuracy 81\%. In \cite{das2015part} Das et al. presented an improved Odia POS tagger using support vector machine, and it achieved 82\% training accuracy. They also used only five tags for manually labelling the dataset having 10k of tokens. Lexicon, NER tags and word suffixes were considered as feature set and fed to the SVM POS tagger. In \cite{ojha2015training} Ojha et al. reported support vector machine (SVM) and conditional random field (CRF) based POS tagger for Odia language achieved an accuracy 91.3\% and 85.5\%, respectively. ILCI dataset with 90k and 2k  tokens was used for training and testing and unigram features were considered during the training period. In this work \cite{pattnaik2020semi}, Pattnaik et al. developed  Hidden Markov Model (HMM) based Odia POS tagger and achieved an accuracy 91.53\% . They implemented the tagger using 0.2 million tokens and 11 tags to annotate the tagset.

	\section{Odia POS Corpus}
	\label{corpus}
In this section, we have provided a brief discussion on the dataset and tagset that were utilised in the development of the Odia POS tagger.

	\subsection{Universal dependency POS tagset for Odia language }
	
	This subsection presents a brief discussion on  UD tagset for Odia part of speech and a mapping from BIS tagset to UD tagset. A tagset contains unique tags which are used to label each word in a text. Universal Dependency (UD) is a framework provides consistent annotation of part of speech across an extensive collection of different human languages \cite{nivre2016universal} \cite{nivre2020universal}. The general idea of UD is to give similar constructions across languages and transfer syntactic knowledge across multiple languages.  The UD annotation approach is founded on an evolution of Google's universal part-of-speech tags \cite{petrov2011universal}, Stanford dependencies \cite{de2008stanford} \cite{de2014universal}, and the Interset interlingua for morphosyntactic tagsets \cite{zeman2008cross}. The ultimate goal of UD is to produce a linguistic representation that will be useful for the study of morphosyntactic structures, the interpretation of semantic meaning, and the application of natural language processing to human languages. UD bears a striking resemblance to the conventional parts of speech. UD is somewhat similar to the conventional parts of speech, and it recognises 17 different classes of words in addition to a variety of other text elements and assigns labels to each of these categories. These categories have a significant amount of linguistic support around the globe.

	Researchers have used different tagsets to develop a POS tagger for the Indian language. Some of are  Ekbal et al. \cite{ekbal2007bengali}  (2007) developed Bengali POS Tagger using 26-tags tagset, Suraksha et al. \cite{suraksha2017part} (2017) presented a Kannada tagger contains 19 tags in tagset, Dhanalakshmi et al. \cite{dhanalakshmi2009tamil} (2009) defined a Tamil POS tagset containing 32-tags,  and Warjiri et al. \cite{warjri2019identification} (2019)introduce a Tagger for Khasi language has 54 labels. The use of different tagsets makes it difficult in the comparative study. To resolve the issue, various tagset schemes are developed. However, since all other datasets of various language around the globe  have been used the corpus annotated with Universal Dependencies tagset in the current experiments. Most of the languages follow the UD tagging framework to maintain uniformity among all languages. However, since there is no availability of Odia dataset annotated with UD POS tagset, it is necessary that Odia dataset also uses the same tagset to maintain uniformity. The collected ILCI corpus was annotated using the Bureau of Indian Standards (BIS) tagset. So, We constructed a simple mapping of tags from BIS tagset to  UD tags based on its respective category. The mapping is given in Table \ref{Table_1}.
	
	%The UD annotation scheme is based on an evolution of Stanford dependencies [][][], Google universal part-of-speech tags [], and the Interset interlingua for morphosyntactic tagsets []. The goal of UD is to offer a linguistic representation that is useful for morphosyntactic research, semantic interpretation, and for practical natural language processing across different human languages. UD offers tokenized sentences with annotations ideal for multi-task learning, including lemmas (LEMMAS), treebank-specific part-of-speech tags (XPOS), universal part-of-speech tags (UPOS), morphological features (UFEATS), and dependency edges and labels (DEPS) for each sentence. UD stays fairly close to traditional parts of speech  UD distinguishes 17 classes of words and other text elements and assigns them the labels (“universal part-of-speech tags,” UPOS). These categories are widely attested in the world’s languages.

	Mapping from Odia POS BIS tagset to UD POS tagset is relatively straightforward and easy to implement. Many of the tags mapped directly from BIS tagset to UD tagset, some of are noun, pronoun, verb, adverb and adjective and so on. A couple of tags made difficult to map because there is no proper class define in UD tagset, like demonstratives, quantifiers and echo-words.  General quantifiers do not indicate any precise quantity and it occurs as intensifier in Odia language. The most common general quantifiers used in Odia language are adhika (much), kebala (only), Ahuri (more) etc. So general quantifiers mapped to intensifier, that is particles in UD tagset. Other two quantifiers in BIS tagset is directly mapped to numeral in UD tagset. BIS tagset do not have determiners and adpositions as a separate category but they have demonstratives and postposition which do not appear in UD. So he label postposition and demonstrative are grouped as adposition and determiner tag in UD tagset respectively. Echo word is one of POS class in BIS tagset but UD POS tagset does not have echo word category. In Odia sentence echo words also used as noun, adjective, adverb etc. depends upon its use in a sentence. From the below examples, we observed that, echo word can have mapped to different POS categories based upon syntactic structure of Odia language. TDIL Odia POS dataset contains 268 echo words. So we manually mapped the echo words to its respective group.

	\begin{figure}[H]
		\centering
		\small
			%\caption*{}
		\includegraphics[width=.7\linewidth]{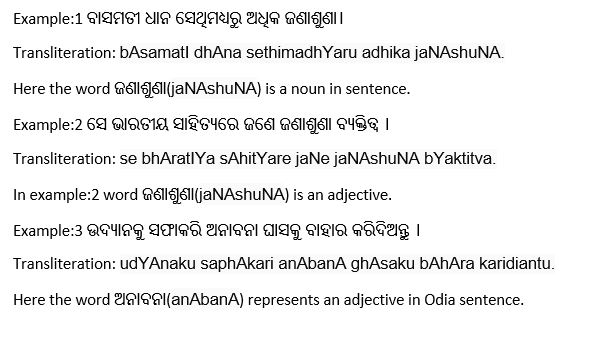}
	
		%\Description{nlp}
	\end{figure}

	%\begin{table}[h]
	%	\centering
	%	%	\setlength{\tabcolsep}{12pt}
	%	\caption{\textbf{UD tagset for Odia language}}
	%	\begin{tabular}{|c|c|c|c|c|}
		%		\hline 
		%		Tag&Description&Example& Transliteration & Meaning\\ \hline 
		%		ADJ&adjective &&&\\
		%		ADP & adposition&&&\\
		%		ADV &adverb&&&\\
		%		AUX&auxiliary&&&\\
		%		CCONJ&coordinating conjunction&&&\\
		%		DET&determiner&&&\\
		%		INTJ&interjection&&&\\
		%		NOUN&noun&&&\\
		%		NUM&numeral&&&\\
		%		PART&particle&&&\\
		%		PRON&pronoun&&&\\
		%		PROPN&proper noun&&&\\
		%		PUNCT&punctuation&&&\\
		%		SCONJ&subordinating conjunction&&&\\
		%		SYM&symbol&&&\\
		%		VERB&verb&&&\\
		%		X&other&&&\\ \hline
		%	\end{tabular}
	%	\label{table_1}
	%\end{table}
	
	\begin{table*}[]
		\centering
		\footnotesize
		\setlength{\tabcolsep}{15pt}
		\caption{\textbf{Mapping from BIS tagset to UD tagset}}
		\begin{tabular}{|c|c|c|c|c|}
			\hline 
			Sl. No.& BIS Tag & BIS Category	 & 	UD Tag & 	 UD Category	 	 	\\
			\hline
			\rownumber			& N\_NN	 &	Common Noun	 & 	  	NOUN &	Noun		 	\\
			\rownumber			& N\_NNV & 	Verbal Noun	&		NOUN &  	Noun	 	 	\\
			\rownumber			& N\_NST	 &	Nloc	& 	NOUN & 	 	Noun	 	 	\\
			\rownumber		& 	N\_NNP	 &	Proper Noun & 	PROPN &  Proper Noun		\\
			\rownumber			& PR\_PRP	 & 	Personal Pronoun  & 	PRON	 & 		Pronoun		 \\
			\rownumber			& PR\_PRF	 & 	Reflexive & 	PRON	 & 		Pronoun	 	 \\
			\rownumber		& PR\_PRL	 & 	Relative Pronoun& 	PRON	 & 		Pronoun	 	\\ 	
			\rownumber		& 	PR\_PRC	 &	Reciprocal	& 	PRON &  	Pronoun	 	 \\
			\rownumber		& PR\_PRQ	 &	Wh-word	& 	PRON & 	 	Pronoun	 	 \\
			\rownumber		& PR\_PRI	 &	Indefinite Pronoun	& 	PRON & 	 	Pronoun	 	 \\
			\rownumber		& 	DM\_DMD &	Deictic Demonstrative	 & 	DET & Determiner		 	\\
			\rownumber		& DM\_DMR &	Relative Demonstrative	& 	DET & 	Determiner	 	 	\\
			\rownumber		& DM\_DMQ &	Wh-word Demonstrative  & 	DET &Determiner		 	\\
			\rownumber		& DM\_DMI &	Indefinite Demonstrative & 	DET &Determiner	 	 \\
			\rownumber		& V\_VM &	Main Verb & 	VERB	 & 	Verb	 	  	\\
			\rownumber		& V\_VM\_VNF &	Finite verb	& 	VERB & 	 Verb	 	  	\\
			\rownumber		& V\_VM\_VNIF &	Non Finite verb	& 	VERB & 	Verb	 	  	\\
			\rownumber		& V\_VM\_VNG &	Gerund	& 	VERB & 	Verb	 	  	\\
			\rownumber			& V\_VAUX &	Auxiliary	 & 	AUX & 	Auxiliary		 \\ 	
			\rownumber			& JJ	 &	Adjective	& 	ADJ	 & 	 	Adjective	  	\\
			\rownumber			& RB &	Adverb	 & 	ADV & 	 Adverb		 	\\
			\rownumber			& PSP &	Postposition & 	ADP & Adposition	 	 \\ 	
			\rownumber			& CC\_CCS &	Subordinating Conjunction & SCONJ & Subordinating Conjunction 	  	\\
			\rownumber			& CC\_CCS\_UT &	Quotative Conjunction & SCONJ &  Subordinating Conjunction 	  	\\
			\rownumber			& CC\_CCD &	Coordinating Conjunction & CCONJ &Coordinating Conjunction 	  	\\
			\rownumber			& RP\_RPD	 &	Default Particle & 	PART	 & 	 	Particle	 	  	\\
			\rownumber		& RP\_INJ	 &	Interjection  & 	INTJ	 & 	 	Interjection		 \\ 	
			\rownumber		& RP\_INTF	 &	Intensifier & 	PART	 & 	 	Particle	 	 \\
			\rownumber		& RP\_CL	 &	Classifier	& 	PART & 	 	Particle	 	 \\
			\rownumber		& 	RP\_NEG &	Negation	& 	PART & Particle	 	  	\\
			\rownumber		& QT\_QTF	 &	General Quantifier	& 	PART & 	 	Particle	 	\\
			\rownumber		& QT\_QTC &	Cardinal Quantifier	& 	NUM & 	Numeral	 	  	\\
			\rownumber		& QT\_QTO &	Ordinal Quantifier	& 	NUM & 	Numeral	 	  	\\
			\rownumber			& RD\_PUNC &	Punctuation	& 	PUNCT & 	Punctuation	 	  \\	
			\rownumber			& 	RD\_SYM &	Symbol	& 	SYM & Symbol	 	 	\\
			\rownumber		& RD\_RDF &	Foreign Word	& 	X & 	Other	 	  	\\
			\rownumber	& RD\_UNK &	Unknown	& 	X & 	Other	 	\\ \hline
			
		\end{tabular}
		\label{Table_1}
	\end{table*}

	\subsection{Corpus Information}
	
	Under the project Indian Languages Corpora Initiative phase-II, funded by the Ministry of Electronics and Information Technology, Government of India, Jawaharlal Nehru University, New Delhi, we accessed two publically accessible corpus via the TDIL website. The first corpus contains monolingual Odia sentences with POS tags, whereas the second corpus contains parallel sentences having POS tags from different domains. Both the corpus having approximately 51,150 sentences. In the corpus, we identified that some special unwanted characters were associated to a number of actual tags. Therefore, the total number of unique tags increased to 87, but only 38 unique tags from the BIS tagset were used to tag the dataset. There were also empty lines, untagged sentences, and redundant space characters in the corpus. Therefore, it is necessary to do data processing prior to utilizing the data to train the model. After data cleansing, we obtained 48,000 Odia POS-tagged sentences. From the available corpus, we have split the dataset to Training, validation and test set according to 70\%, 15\%, and 15\% respectively.	Deatails of the dataset shown in Table \ref{Table_2}. The distribution of the ILCI corpus over the UD tagset is presented in reference Table \ref{Table_3}.

	\begin{table}[h]
		\centering
		\caption{\textbf{Odia POS corpus details}}
		\begin{tabular}{|c|c|c|c|}
			\hline 
			
			Collected Corpus &Domain &NO. of Sentences& NO. of tokens \\ \hline
			Odia Monolingual Text Corpus ILCI-II & Agriculture, Entertainment &19,000 &2,76,916 \\ \hline
			Hindi-Odia Text Corpus ILCI-II &Agriculture, Entertainment 
			&31,500 &3,94,438 \\ \hline

		\end{tabular}
		\label{Table_2}
	\end{table}

	\begin{table}[h]
		\centering
		\setlength{\tabcolsep}{15pt}
		\caption{\textbf{Distribution of ILCI Odia POS corpus over UD tagset}}
		\begin{tabular}{|c|c|c|}
			\hline 
			Tag&Description&Count\\ \hline 
			NOUN&noun&258726\\
			VERB&verb&86369\\
			PUNCT&punctuation&68322\\
			PROPN&proper noun&58866\\
			ADJ&adjective &43502\\
			NUM&numeral&28904\\
			CCONJ&coordinating conjunction&26170\\
			PART&particle&26048\\
			PRON&pronoun&22426\\
			DET&determiner&19337\\
			ADP & adposition&13255\\
			ADV &adverb&9387\\
			SCONJ&subordinating conjunction&6485\\
			X&other&1571\\ 
			AUX&auxiliary&1397\\
			INTJ&interjection&444\\
			SYM&symbol&145\\ \hline
		
		\end{tabular}
		\label{Table_3}
	\end{table} 

	\section{System Model}
	\label{model}
	The statistical and deep learning models that were utilised to construct the Odia POS tagger have been explained in this section of the article.

	\subsection{Odia POS Tagger developed using statistical method}
	
	Several existing models are present for sequence labelling tasks like POS tagging, NER and chunking etc. HMM, MEMM and CRF are the stochastic methods used to solve sequence labelling problems. However, from the literature survey, we knew that conditional random field performs better than HMM and MEMM. CRF model is the most commonly used sequencing model for labelling tasks. Moreover, the CRF model makes it easy to examine the word before and after entity by its undirected graph property.
	CRFs are the type of discriminative model that learns the conditional probability distribution. Conditional random fields is an undirected graph-based method where node represents input or observation sequence x corresponds to label, or output sequence y. This model aims to find the output label or tag y that maximises the conditional probability P(y|x) for a given input sequence x.
	For finding the most probable sequence label from the given word sequence, mathematically, we can write it as Y = argmax P(x|y).	Let the observation word sequence $x= (x_1x_2x_3…..x_n)$
	and corresponding tag sequence $y=(y_1y_2y_3……….y_n)$.
	
	Mathematically, The conditional probability chain structure of the crf model can be expressed as 
	
	\begin{equation}
		P(y \mid x)=\frac{1}{Z_x}  \exp \left(\sum_{t=1}^n \sum_{k=1}^K \lambda_{k} \cdot f_{k}\left(y_{t-1}, y_{t}, x\right)\right)
	\end{equation}
	
	Where $ \lambda $ is the weight associated with each distinct features K, and the model depends on the set of features. $f_{k}\left(y_{t-1}, y_{t}, x\right)$ denotes feature function whose value is represented in binary. It may be ‘0’ or ‘1’.
	$ Z_x $ is used for normalization and sum of probability of all state sequences is one.
	$ Z_x $ can be represented as 
	
	\begin{equation}
		Z(x)=\sum_{y} \exp \left(\sum_{t=1}^n \sum_{k=1}^K \lambda_{k} \cdot f_{k}\left(y_{t-1}, y_{t}, x\right)\right)
	\end{equation}
	
	All the details of feature set, we have explained in the result section.
	
	\subsection{Odia POS Tagger developed using deep learning approach}
	
%	In this subsection, we constructed a POS tagger for the Odia language using various deep learning-based models, including CNN, Bi-LSTM, and a combination of CNN and Bi-LSTM with CRF. Figure \ref{deep_learning} illustarates the architecture of deep learning based model used for Odia POS tagging. For the sake of simplicity, we summarized the steps involved in creating the deep learning model based on POS tagging for the Odia language. The overall architecture Bi-LSTM model used for Odia POS tagging is shown in Figure \ref{lstm1}.

\begin{figure}[h]
		\centering
		\large
		\includegraphics[width=.8\linewidth]{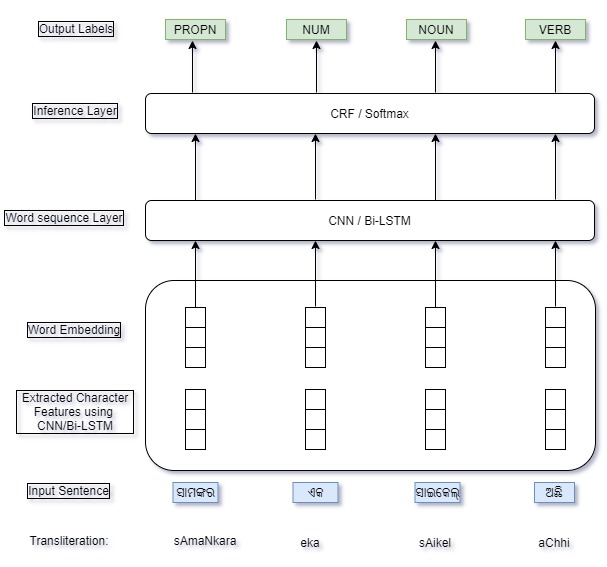}
		\caption{Architecture of Odia POS tagger using deep learning-based model}
		%\Description{nlp}
		\label{deep_learning}
	\end{figure}

\begin{figure}[h]
		\centering
		\large
		\includegraphics[width=0.95\linewidth]{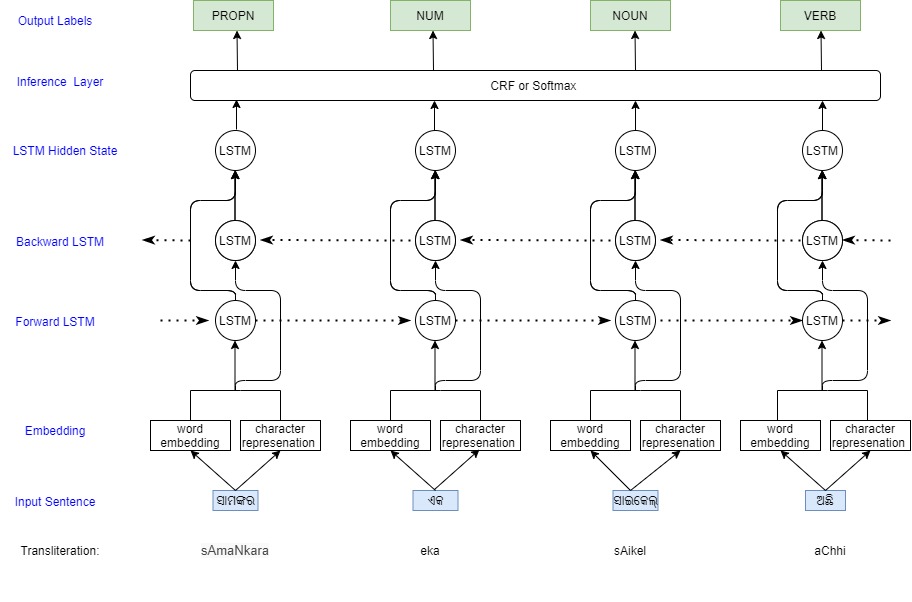}
		\caption{Architecture of Odia POS tagger using Bi-LSTM model}
		%\Description{nlp}
		\label{lstm1}
	\end{figure}

In this subsection, we discuss the development of Odia POS tagger using deep learning based approaches. Here, we built a POS tagger for the Odia language by employing a variety of deep learning-based models, such as CNN, Bi-LSTM, and a combination of CNN and Bi-LSTM with CRF. The architecture of the deep learning-based model used for Odia POS tagging is shown in Figure \ref{deep_learning}. For the sake of simplicity, we have summarised the steps involved in developing the deep learning model for Odia based on POS tagging below. Figure \ref{lstm1} illustrates the overall architecture of the Bi-LSTM model used for Odia POS tagging.
	
	\begin{enumerate}
		
		\item An Odia sentence is taken as an input to the model.
		\item CNN or Bi-LSTM neural encoders are used to integrate character sequence information of Odia word as character-level embedding.
		\item We generated a word embedding vector of more than 4.5 million Odia sentences from different resources as initialization for word-level embedding using fastText word embedding method.
		\item We obtained character level embedding and word-level embedding are concatenated and fed into a fully connected neural network.
		\item The output of the previous step is taken input to the word sequence layer, and final word embedding is fed into the word sequence layer.
		\item Finally, last hidden layer's output of the word sequence layer was taken input to the inference layer (softmax or CRF) to predict the possible tags for each input sentence.
	\end{enumerate}
	
	\subsubsection{\textbf{Character sequence layer}}

	For sequence labelling tasks, it is important to consider word morphology. Character-level embedding is taken to represent characters of input words. It is used to deal with language with complex morphology. Character embeddings can be used to represent character features, such as prefixes and suffixes, among other character features. In order to extract the character-level information, we made use of two different kinds of networks, CNN and Bi-LSTM.
	% CNN is frequently considered an example of a neural network that incorporates character-level representation. CNNs are used primarily in image processing, but more recently they have also found widespread application in natural language processing (NLP).%
	 The CNN approach is an effective method for extracting morphological information of characters from words that are given. Each character within a word token is initially mapped to a character vector. Then, filters of varying sizes are applied to the embedding matrix in order to capture the key characteristics of nearby inputs. The final stage of CNN consists of an operation called max-pooling, which extracts a single feature from each of the feature maps.  After that, the output characteristics are concatenated in order to maintain the information that is specific to each word. Moreover LSTM models are also applied for character-level information extraction. Bidirectional LSTM or Bi-LSTM is used to record sequence information from left to right (forward LSTM) and right to left (backword LSTM). Using bidirectional hidden states, the model can then preserve both past and future knowledge. In addition, the Bi-LSTM model captures the global characteristics of each word token. Then, concatenate the two final hidden states of the forward and backward LSTM to get a vector of fixed size representing a word token.

	\subsubsection{\textbf{Word Embedding}}

	In many NLP tasks, nearby words play a crucial role. In addition, the performance of sequence labelling tasks, such as POS tagging, NER, and Chunking, is dependent on the surrounding terms. In order to accomplish this, we require word embedding for distributed word representation, which maps words into a low-dimensional vector space. Word vector has the advantage of capturing  relationships between words. NLP offers a variety of strategies for word embedding. In our work, we used fastText \cite{bojanowski2017enriching}, one of the embedding techniques developed by Facebook. We train the fastText model because each word is represented as an n-gram of characters, it preserves subword information, and it can compute valid word embeddings for out-of-vocabulary terms. Therefore, it can provide the vector for unseen words during word embedding training. The word vectors are generated after training the fastText model. To learn word representations, contextually related words often occur with similar surrounding words. However, there are no publicly available pre-trained Odia word vectors. Therefore, we choose to train the models with the fastText model. For training purposes, a huge corpus of Odia text is required. Therefore, we gathered raw corpora from number of resources \cite{parida2020odiencorp} \cite{parida2020odiencorp2} \cite{ramesh2022samanantar} \cite{kunchukuttan2020indicnlpcorpus} and Odia-specific websites. The collection includes around 4.5 million Odia sentences. The input for fastText training is a large corpus of text, and the output is a vector with several hundred dimensions for each unique word in the corpus. The model contains two network-based variants: Continuous Bag of Words (CBOW), and Skip-gram. In this work, the skip-gram approach was chosen for the training process. The input for the Skip-gram architecture is a single word, and it generates predictions for words that are contextually connected to that word. Run time and the quality of the trained model are both impacted by a number of different parameter selections. After conducting the experiment with a number of different vector dimension values (ranging 100, 150, 200, 250, and 300), we observed that value of 200 worked well in our experiment. We also experimented with different window sizes and determined the best one to be 5. The word vectors are generated after training the fastText model.

	\subsubsection{\textbf{Word sequence layer}}
	%Like character sequence layer word sequence layer encompasses both CNN and Bi-LSTM neural network.
	%Biderectional LSTM are captured arbitrary long context information to overcome the limitations of a fixed window size.
	%Input of word sequence layer is the word representation,which is the combination of word embedding and character sequence representation. For many sequence labeling tasks it is benifial to have acess of both past and future contextual information. However simple LSTM networks takes information from only past,having no idea about future context.  Bidictional LSTM are able to capture the left and right contextual information of each word. Two separe hidden states of BiLSTM to catures the past and future information and then concatenated to form the final output.
	Similarly to the character sequence layer, the word sequence layer consists of both CNN and Bi-LSTM neural networks. Word representations, which may comprise word embeddings and character sequence representations, are the input of the word sequence layer. Stacking the word sequence layer enables the development of a more robust feature extractor. CNN uses a sliding window to collect local characteristics and a max-pooling to encode the aggregated character sequence. Immediately following each CNN layer, batch normalization and dropout are applied.
	
	Bi-LSTM, one of the types of recurrent neural network, was also employed at the word sequence layer (RNN). In order to overcome the limits of a set window size, a bidirectional LSTM captures arbitrary length context data. It is helpful to have access to knowledge about the sequence's context in the past as well as the sequence's context in the future for many sequence labelling tasks. On the other hand, simple LSTM networks merely take input from the past and are unaware of the context of the future. Bi-LSTM are capable of remembering both the left and right contextual information associated with each word. Two distinct hidden states of the Bi-LSTM are used to capture information from the past and the future, which is then concatenated together to produce the final output.
	
	\subsubsection{\textbf{Inference Layer}}
	%This layer is the last layer of our BiLSTM deep learning-based approach for Odia POS tagging. Here the output of BiLSTM ntwork are input to the inference layer and  inference layer assign labels to each words in the sentence or document. In this, we applied CRF and softmax as the output layer. [] reported that CRF in output layer boost the performance of sequence labelling task. CRF layer learns from the consecutive labels by using state transition matrix as parameters, by which we can predict the current tag. In some sequence labelling tasks, softmax performs better than CRF due to the support of parallel decoding[].
	
	This is the final layer of our deep learning-based strategy for POS tagging in Odia. Here, the output of the word sequence layer is input to the inference layer, which assigns labels to each word in the sentence or document. In this, we applied CRF and softmax as the output layer. We found in the literature that CRF in the output layer improves the performance of some sequence labelling tasks. CRF layer acquires knowledge from successive labels by employing state transition matrix as a set of parameters; with this information, we are able to anticipate the current tag. Because it allows for concurrent decoding, softmax performs better than CRF in certain sequence labelling tasks \cite{ling2015finding}. 
	
	\section{Experimental result}
	\label{result}
	\subsection{CRF model result}
	%
	%       In this subsection, we briefly presented training and testing using CRF model and defined a suitable feature set to interpret their influence in Odia POS system. Here we observed the effectiveness of CRF model for Odia POS tagging using UD tagset on TDIL data. LBFGS (Limited-Memory Broyden-Fletcher-Goldfarb-Shanno) is a quasi-newton algorithm used as an optimizer to train our CRF model. For the implementation, we used CRF++-0.58.8. We have used the default parameter settings of the tool except c = 1.5 and f = 3. Here, 'c' parameter trades the balance between over-fitting and under-fitting, and the 'f' parameter sets the cut-off threshold for the features. These values gave the best result in our experiments.
	
	In this subsection, we provided a brief description of training and testing using the CRF model and established a relevant feature set for interpreting their impact on the Odia POS system. We investigated the efficacy of our approach using the ILCI Odia dataset. Using ILCI Odia corpus, we evaluated the performance of the CRF model for Odia POS tagging with the BIS and UD tagsets. Our CRF model is trained using LBFGS (Limited-Memory Broyden-Fletcher-Goldfarb-Shanno), a quasi-Newton approach and this algorithm is used for large-scale numerical optimization problems. We used CRF++-0.58.8 toolkit to prepare the model and the toolkit runs with the CRF algorithm. For our experiment, we first defined the feature set to train the model using the feature templet of the CRF++ toolkit. To improve the overall performance of any machine learning task, it is necessary to have the feature set, which should include valuable features. Feature templet contains unigram features to define contextual information during training. A prefix, an identifying number, and a rule string are included in each template. The template type can be determined by the prefix; for example, "U" denotes a unigram template. An identification number is assigned to each template so that they may be distinguished from one another, and a rule string is utilised to guide CRF in the generation of features.
	
	\begin{figure}[h]
		\centering
		
		\large
		\includegraphics[width=0.8\linewidth]{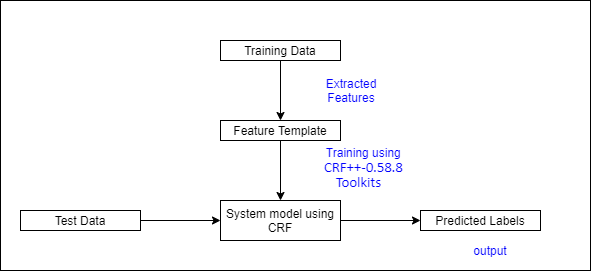}
		\caption{Flow Chart for CRF approach}
		\Description{nlp}
		\label{crf_fig}
	\end{figure}      
	The information regarding the previous word, the next word, the current word, as well as the tag of the current word, is contained within the defined feature templates. According to the research, information about the root word and its morphological inflections is quite helpful for the POS labelling task. The suffix and prefix information was then added to the feature set. For the development of the CRF classifier, a number of experiments were conducted using various combinations of individual features to identify the feature set with the highest accuracy. In Table 4, we have provided a summary of the many possible feature combinations that were utilised in our experiment. In the construction of the word "features," a variety of different combinations of preceding and following words were utilised. In addition, we made use of suffix and prefix information, with the length of the suffix varying between three, four, and five characters and the length of the prefix varying between two and three characters. Figure \ref{crf_fig} presents the flow chart of POS tagger using CRF++-0.58.8 toolkit.	
	
	CRF models are trained using the regularisation technique L2 to generate all features needed to calculate probabilities at the time of tagging. We trained the model multiple times by modifying the parameter, and the optimal result was obtained by setting the -c and -f parameters to 1.5 and 2, respectively. Here, 'f' defines the cut-off threshold for the features, whereas 'c' specifies the balance between over-fitting and under-fitting. We evaluated the efficacy of the CRF tagger using approximately 2,76,900 tokens from Corpus I and about 6,711,000 tokens from combined Corpora I and II on BIS and UD tagsets. In Tabel \ref{Table_4}, we summarized the combination of features represented by feature id and corresponding accuracy. Here we can observe from Table \ref{Table_5} that the accuracy increases as the corpus size grows. In these experiments, CRF tagger with defined feature sets obtained a maximum accuracy of 92.08\%.      

	\begin{table}[h]
		\centering
		\caption{\textbf{Description of feature set}}
		\begin{tabular}{|c|c|c|}
			\hline 
			Feature ID&Feature Template&Description\\ \hline 
			$f_1$	& "U01:\%x[-1,0]\%x[0,0]\%x[1,0]" &Current word + Previous word + Next word \\ 
			$f_2$	& "U02:\%x[-2,0]\%x[-1,0]\%x[0,0]" & Current word + Previous two words \\ 
			$f_3$	& "U03:\%x[0,0]\%x[1,0]\%x[2,0]" &Current word + Next two words \\ 
			
			$f_4$	&"U04:\%x[0,1]" &POS Tag of Current word \\ 
			$f_5	$& "U05:\%x[0,2]" & Suffix length 3 of current word\\ 
			$f_6$	& "U06:\%x[0,3]"&Suffix length 4 of current word \\ 
			$f_7$	& "U07:\%x[0,4]" &Suffix length 5 of current word \\ 
			$f_8$& "U08:\%x[0,5]" &Prefix length 2 of current word \\
			$f_9$& "U09:\%x[0,6]" &Prefix length 3 of current word \\  \hline 
		\end{tabular}
		\label{Table_4}
	\end{table}
	
	\begin{table}[h]
		\centering
		\large
		\caption{\textbf{Accuracy of CRF based tagger using different features}}
		\begin{tabular}{|c|c|c|A|A|}
			\hline 
			Sl No. & Dataset& Feature details &Test accuracy on BIS tagset &Test accuracy on UD tagset \\ \hline 
			1&  & $f_1$ &87.56 & 89.24\\
			2& & $f_2$&87.02 &88.55\\
			3& 1,93,840 tokens for training & $f_3$ & 87.33 & 88.82 \\
			4& 41,537 tokens for training & $f_1 + f_4$ &87.74 &89.49 \\
			5&  & $f_1 + f_4 + f_5 + f_6 + f_7$  & 87.98& 89.65\\
			6& & $f_1 + f_4 + f_5 + f_6 + f_7 + f_8 + f_9 $& 88.14& 89.97\\ \hline
			1&  & $f_1$ & 89.76&91.42 \\
			2& & $f_2$ & 88.87&90.15\\
			3&  4,69,950 tokens for training& $f_3$ & 89.01&90.59\\
			4& 1,00,704 tokens for testing & $f_1 + f_4$ & 89.77 &91.50\\
			5&  & $f_1 + f_4 + f_5 + f_6 + f_7$  & 90.41&91.83 \\
			6& & $f_1 + f_4 + f_5 + f_6 + f_7 + f_8 + f_9$ &90.45 &92.08\\ \hline
			
		\end{tabular}
		\label{Table_5}
	\end{table}

	\subsection{Deep learning model results}

	This subsection provides a brief explanation of the experimental outcomes of deep learning-based technique for the Odia POS tagger. We used the same corpus for this experiment as we did in our initial CRF-model-based approach. Approximately 2,76,900 tagged tokens (36505 unique words) and 6,71,000 tagged tokens (75394 unique words) are fed into the systems. For our implementation, 70\% of the corpus was used for training, 15\% for validation, and the remaining 15\% was used for testing. For our experiment, we employed CNN and Bi-LSTM, two deep learning-based approaches, to construct a POS tagger for the Odia language. The training for both of these models was done with the use of hyper-parameter values that gave the best results on the validation set. The selected hyper-parameters that were used in our experiments are summarised in Table \ref{Table_6}. To initialize the word, we used the pre-trained word embedding dimension of 200 and the character embedding size of 30. The stochastic gradient descent (SGD) optimizer is applied in order to optimize the parameters with a minimum batch size of 10. The Bi-LSTM and CNN models were trained using 50 and 100 epochs, respectively, and their respective learning rates were 0.015 and 0.005. In addition to this, we used the regularization approaches of l2 regularization and a dropout value of 0.3 in order to enhance the performance of the model on the unseen data.
	
	To estimate robustness of models with respect to features, we train on six different combination on both of the model including Bi-LSTM, CNN, WE + Bi-LSTM, WE + CNN,  WE + CharCNN + Bi-LSTM, WE + CharCNN + CNN, WE + CharBi-LSTM + Bi-LSTM, WE + CharBi-LSTM + CNN, WE + CharCNN + Bi-LSTM + CRF, WE + CharCNN + CNN + CRF, WE + CharBi-LSTM + Bi-LSTM + CRF, and WE + CharBi-LSTM + CNN + CRF. Both corpora are used to train every model. There are three CRF-based models and three Softmax-based models with distinct representations of character and word sequences. Here token accuracy is used to evaluate the system performance. The results of Odia POS tagger on BIS and UD tagset using the deep learning-based technique are presented in Table \ref{Table_7}. In Table \ref{Table_7}, CharCNN indicates models that use CNN for character sequence information, CharBi-LSTM indicates models that use Bi-LSTM for character sequence information, and WE indicate word embedding information. Among the various combination of features, the results demonstrate that the Bi-LSTM with pre-trained word vector and character information using CNN provided the highest accuracy of 94.58 percent. Also, we observed that models with character-level information perform better than models without character-level information. It is found that the accuracy of deep learning-based approaches for Odia POS tagger on BIS and UD tagsets outperforms the accuracy of existing work.
	
	\begin{table}[h]
	\centering

	\caption{\textbf{Hyper-parameters for deep learning model}}
	\begin{tabular}{|c|c|}
		\hline 
		Parameters&value\\ \hline 
		Character embedding dimension&30\\
		Hidden layer character dimension& 50\\
		Word embedding dimension & 200\\
		Hidden layer word dimension & 300\\
		Optimizer & SGD\\
		Learning rate for Bi-LSTM model& 0.015\\
		Learning rate for CNN model& 0.005\\
		Learning decay rate & 0.05\\
		Dropout& 0.3\\
		Batch size& 10\\
		Number of epochs for Bi-LSTM model& 50\\
		Number of epochs for CNN model& 100\\
		\hline
	\end{tabular}
	\label{Table_6}
\end{table}

	\begin{table*}[]
		\centering
		\caption{\textbf{Comparsion of tagging perfermance on POS tasks for various neural network model}}
		\begin{tabular}{|c|c|A|A|}
			\hline 
			Dataset  & Model &Test accuracy on BIS tagset&Test accuracy on UD tagset\\ \hline 
			
			\multirow{ 6}{*}{ }& 	CNN &87.47&88.85 \\
			&WE +CNN &88.72 &90.14\\
			1,93,840 tokens for training		&WE + CharCNN + CNN  &89.31 &90.81 \\
			41,537 tokens for validation		&WE + CharBi-LSTM + CNN  &89.13 &90.67 \\
			&WE + CharCNN + CNN  + CRF  & 89.01& 90.44 \\
			&WE + CharBi-LSTM + CNN  + CRF  &88.91&90.34\\ \hline
			
			\multirow{ 6}{*}{}	& 	CNN &89.32&91.02\\
			&WE +CNN &90.74 &92.31\\
			4,69,950 tokens for training 	&WE + CharCNN + CNN  &91.29 & \textbf{92.77} \\
			1,00,704 tokens for validation	&WE + CharBi-LSTM + CNN  &91.19&92.60\\
			&WE + CharCNN + CNN  + CRF  &91.02& 92.43 \\
			&WE + CharBi-LSTM + CNN  + CRF  &90.89 & 92.35\\ \hline
			\multirow{ 6}{*}{ }& 	Bi-LSTM &88.97&90.24\\
			&WE +Bi-LSTM &90.14 &91.65\\
			1,93,840 tokens for training		&WE + CharCNN + Bi-LSTM  &90.84 &92.15\\
			41,537 tokens for validation	&WE + CharBi-LSTM + Bi-LSTM  &90.65&92.03\\
			&WE + CharCNN + Bi-LSTM  + CRF  &90.49&91.98  \\
			&WE + CharBi-LSTM + Bi-LSTM  + CRF  &90.32&91.83\\ \hline
			
			\multirow{ 6}{*}{ }	& 	Bi-LSTM &90.75&92.65\\
			&WE +Bi-LSTM &92.06 &94.02\\
			4,69,950 tokens for training &WE + CharCNN + Bi-LSTM  & 92.60&\textbf{94.48}\\
			1,00,704 tokens for validation &WE + CharBi-LSTM + Bi-LSTM  &92.44&94.23\\
			&WE + CharCNN + Bi-LSTM  + CRF  &92.32& 94.17 \\
			&WE + CharBi-LSTM + Bi-LSTM  + CRF  &92.21&94.09\\ \hline
			
		\end{tabular}
		\label{Table_7}
	\end{table*}
	
	%\begin{table}[H]
	%	\centering
	%	\large
	%	%	\setlength{\tabcolsep}{12pt}
	%	\caption{\textbf{Comparsion of tagging perfermance on POS tasks for various neural network model}}
	%	\begin{tabular}{|c|c|A|A|}
		%		\hline 
		%		Dataset  & Model &Accuracy on BIS Tagset&Accuracy on UD Tagset\\ \hline 
		%		
		%		 	\multirow{ 6}{*}{ }& 	BiLSTM &88.97&90.24\\
		%		 			&WE +BiLSTM &90.14 &91.65\\
		%		 	2,21526 tokens for training		&WE + CharCNN + BiLSTM  &90.84 &92.15\\
		%		 		55,390 tokens for testing	&WE + CharBiLSTM + BiLSTM  &90.65&92.03\\
		%		 			&WE + CharCNN + BiLSTM  + CRF  &90.49&91.98  \\
		%		 			&WE + CharBiLSTM + BiLSTM  + CRF  &90.32&91.83\\ \hline
		%		
		%	\multirow{ 6}{*}{ }	& 	BiLSTM &90.75&92.65\\
		%	&WE +BiLSTM &92.06 &94.02\\
		%	5,36,350 tokens for training &WE + CharCNN + BiLSTM  & 92.60&94.58\\
		%	1,35,004 tokens for testing &WE + CharBiLSTM + BiLSTM  &92.44&94.23\\
		%	&WE + CharCNN + BiLSTM  + CRF  &92.32& 94.17 \\
		%	&WE + CharBiLSTM + BiLSTM  + CRF  &92.21&94.09\\ \hline
		%
		%	
		%	\end{tabular}
	%	\label{table_1}
	%\end{table}

	\subsection{Comparison with existing POS Tagger in Odia Language}
	
	We compared the performance of our CRF, CNN and Bi-LSTM model with the results of previous work on Odia POS tagging. Table \ref{Table_8} presents the results of other researchers and the results of our model. From the other researchers' results, we can see that Pattnaik et al. \cite{pattnaik2020semi} presented Odia POS tagger on one of statistical model for sequence labelling task, that is HMM model gives the highest accuracy 91.53. On the other hand, the proposed POS tagger on the CRF model gives 92.08\% accuracy using surrounding word information prefix and suffix information features to increase the accuracy. We observed from Table \ref{Table_8}, result achieved by the CNN and Bi-LSTM model using BIS and UD tagsets on ILCI corpus gives better accuracy compared to the existing research work.
	
	\begin{table}[h]
		\centering
		\caption{\textbf{Comparsion of tagging perfermance on POS tasks for various neural network model}}
		\begin{tabular}{|c|c|c|c|c|}
			\hline 
			& Model &Dataset&tagset&Accuracy\\ \hline 
			
			\multirow{ 4}{*}{Previous works}& 	
			ANN \cite{das2014novel} & NA & 5tags & 81\%\\
			& SVM  \cite{das2015part} & 10,000 & 5 tags & 82\%\\
			&CRF  \cite{ojha2015training} & training:90k,validation:2k& BIS tagset (38 tags) &82-86\% \\
			& HMM  \cite{pattnaik2020semi} & 0.2 million tokens & 11 tags &91.53 \\ \hline
			
			\multirow{ 4}{*}{Our results }	
			& CRF & Training:4,69,950 ,validation:1,00,704 & UD tagset (17 tags)& 92.08\%\\
			& CRF & Training:4,69,950 ,validation:1,00,704 & BIS tagset (38 tags)& 90.45\%\\
			& CNN & Training:4,69,950 ,validation:1,00,704  &UD tagset (17 tags)&92.74\%\\
			& CNN & Training:4,69,950 ,validation:1,00,704  & BIS tagset (38 tags)&91.29\%\\
			& Bi-LSTM & Training:4,69,950 ,validation:1,00,704  &UD tagset (17 tags)&94.48\%\\
			& Bi-LSTM &Training:4,69,950 ,validation:1,00,704  &BIS tagset (38 tags)&92.60\%\\ \hline

		\end{tabular}
		\label{Table_8}
	\end{table}

	\section{Error Analysis }
	\label{analysis}

	\begin{figure}[h]
		\centering
		\large
		\includegraphics[width=1.1\linewidth]{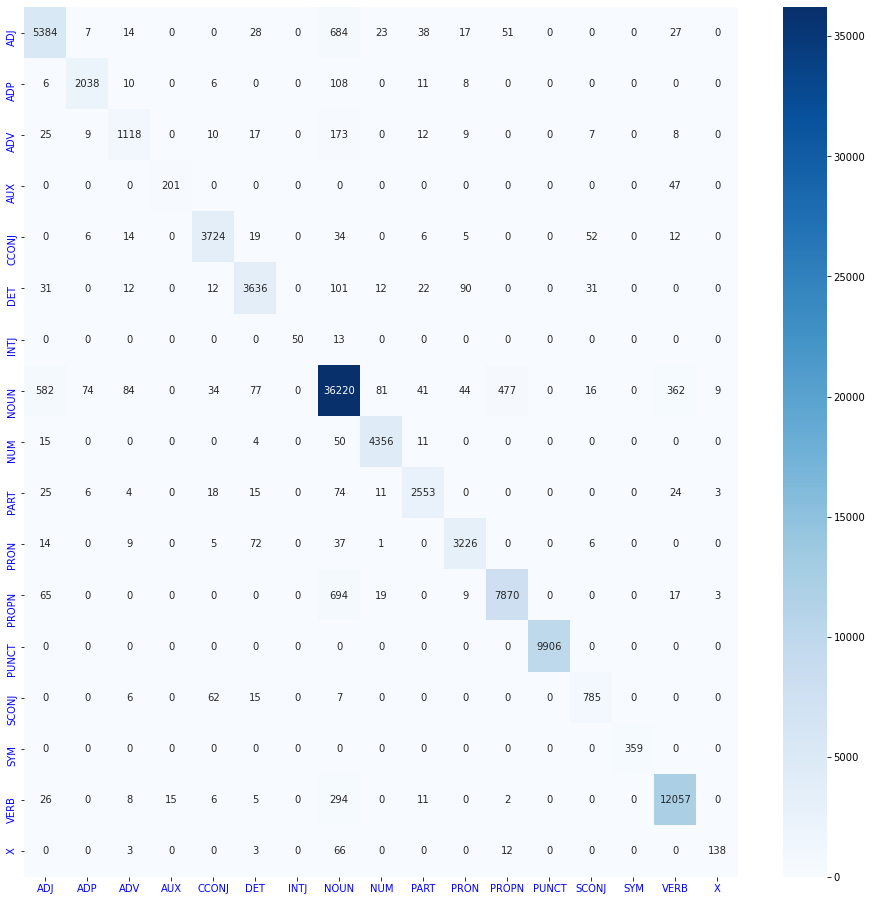}
		\caption{Confusion matrix of Bi-LSTM model}
		\label{bi_lstm_fig}
		%\Description{nlp}
	\end{figure}

	\begin{figure}[h]
		\centering
		\large
		\includegraphics[width=1.1\linewidth]{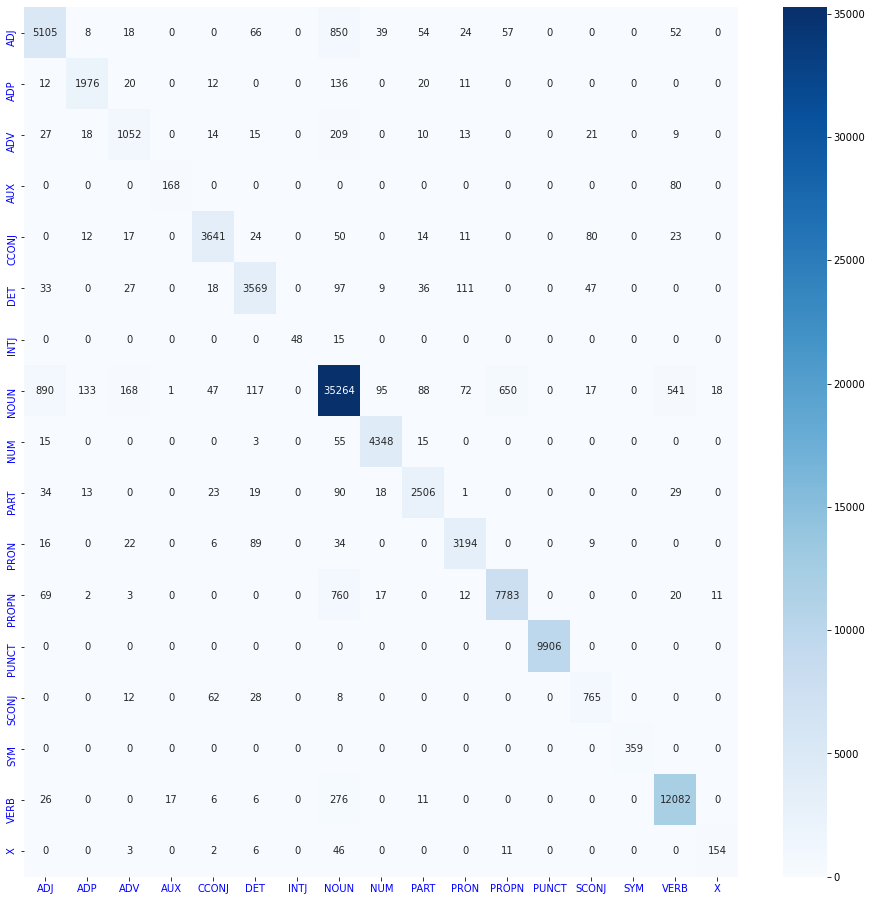}
		\caption{Confusion matrix of CNN model}
		\label{cnn_fig}
	%	\Description{nlp}
	\end{figure}

	An examination of the nature of improperly classified tags is essential to POS tagging. We therefore performed an error analysis on the predicted POS tags using the CNN and Bi-LSTM models. Therefore, we conducted an error analysis for the predicted POS tags by the CNN and Bi-LSTM models. Figure \ref{bi_lstm_fig} and Figure \ref{cnn_fig} presents the confusion matrix of Bi-LSTM and CNN for Odia POS tagger respectively. Table \ref{table_9} presents the misclassification classes for CNN and Bi-LSTM models derived from the respective confusion matrices. Here, we considered the top 5 misclassification labels and calculated their number of instances and error rate.  
	
	\begin{table}[h]
		\centering
		\caption{\textbf{Top five misclassification classes for CNN and Bi-LSTM model}}
		\begin{tabular}{|c|c|c|c|c|}
			\hline 
			Model&Actual label&Predicted label&Number of Instances&Error rate\\ \hline 
			\multirow{ 5}{*}{CNN }&
			NOUN & ADJ&890&12.42\\
			&ADJ & NOUN &850&11.86\\
			&PROPN & NOUN&760&10.61\\
			&NOUN & PROPN&650&9.07\\
			&NOUN & VERB &541&7.55
			\\		 \hline
			
			\multirow{ 5}{*}{Bi-LSTM }&
			PROPN & NOUN&694&12.71\\
			&ADJ & NOUN &684&12.52\\
			&NOUN & ADP&582&10.65\\
			&NOUN & PROPN&477&8.73\\
			&NOUN & VERB &362&6.63
			\\		 \hline
		\end{tabular}
		\label{table_9}
	\end{table} 
	
	From the confusion matrices,we observed that Bi-LSTM model predicts the label more accurately for the majority of classes than the CNN model, whereas the CNN model predicts the label more accurately than Bi-LSTM for some classes. We determined from our experiment that both models predict the same label with a 96.14\% of accuracy on the validation set. The word for which Bi-LSTM and CNN model disagree on the predicted tag, Bi-LSTM correctly labels 63.68\% of the words whereas CNN correctly labels 29.97\% accuracy of the words. 
	%whereas they provided different output tags with a 63.68\% of accuracy for Bi-LSTM model and 29.97\% accuracy for CNN model on test data.
	Therefore, we took into consideration the label information as a feature in which the models give the same predicted tags, and we trained the model with this additional information using Bi-LSTM and CNN in order to improve the performance of the Odia POS tagger. Here we concatenated additional tag information of training and validation data with word embedding vector and the updated word embedding vector is input the Bi-LSTM and CNN model. Let $ Z_w $ be the original word embedding produced by fastText for word w. We created 18 different one hot encoded vectors $ x_0 ,x_1,...x_{17} $. Here $x_1 , x_2,...x_{17}$ represents each pos tags. For word w , if both Bi-LSTM and CNN agreed on the tags (say,noun), then we replaced $Z_w$ with $Z_w \oplus x_1 $. If for word w, they disagree, we replaced $Z_w$ with $ Z_w \oplus x_0$ and also we concatenated $x_0$ for any new words came in vocabulary. Here $ x_0$ represents 0 0 0...0. This modified word embedding was input to the Bi-LSTM and CNN models. We observed the results across five iterations and found that in the third iteration, the validation accuracy was 94.56\% and its corresponding the test accuracy was 94.58\%. Which was the best result among the five iterations and improved over the original accuracy of 94.48\%.  
	 We repeated this procedure and monitored model accuracy until model accuracy improved with each iteration. The result is presented in Table \ref{tabel_10}. We determined from the above table that the Bi-LSTM model achieved the maximum accuracy, 94.58 percent. Hence, based on the above mentioned observations, the experimental results indicate that the Bi-LSTM model with various combinations of features achieved better results for Odia POS tagging.

	%For error analysis, un-tagged words are tested using our BiLSTM system. Here some of the ambiguities are observed. For example, categorial ambiguity arises when a particular word form can, in different instances, represent different grammatical categories. The ambiguity also occurs when a specific form of the word has different tags in the same context. For example, the word "mAne" was tagged as coordinating conjunction, noun or adposition. Word "prathama" was labelled as adverb, noun or numeral. Therefore, words are tagged wrongly by the system due to ambiguity problems. Tagger generated the most common errors because the system is confused with 'noun' with 'adjective', coordinating conjunction with numeral, adverb with adjective, numeral with adjective, adposition with determiner, adpostion with adjective. Table 10 presents some untagged input sentences to the system and gives tagged words as output. 

	\begin{table}[h]
		\caption{\textbf{Updated results of CNN and Bi-LSTM model}}
		\begin{tabular}{|l|l|ll|ll|}
			\toprule
			\multicolumn{1}{|c|}{\multirow{3}{*}{Iteration}} & \multirow{3}{*}{\begin{tabular}[c]{@{}l@{}}Number of   instances\\  result of CNN equals\\  to Bi-LSTM\end{tabular}} & \multicolumn{2}{c|}{\multirow{2}{*}{\begin{tabular}[c]{@{}c@{}}Accuracy of  \\  CNN model\end{tabular}}} & \multicolumn{2}{c|}{\multirow{2}{*}{\begin{tabular}[c]{@{}c@{}}Accuracy of \\   Bi-LSTM model\end{tabular}}} \\
			\multicolumn{1}{|c|}{}                           &                                                                                                                      & \multicolumn{2}{c|}{}                                                                                    & \multicolumn{2}{c|}{}                                                                                        \\ \cmidrule(l){3-6} 
			\multicolumn{1}{|c|}{}                           &                                                                                                                      & \multicolumn{1}{l|}{Validation}                                 & Testing                                & \multicolumn{1}{l|}{Validation}                                   & Testing                                  \\ \midrule
			- &                                                                                                      - & \multicolumn{1}{l|}{92.61}                                           &  92.77      & \multicolumn{1}{l|}{94.43}                                      &    94.48  \\
			
			1 &                                                                                                                    
			544533   & \multicolumn{1}{l|}{92.75}                                           &  92.89   & \multicolumn{1}{l|}{94.48}                                             &  94.54  \\
			
			2 &                                                                                                                      
			549850 & \multicolumn{1}{l|}{93.03}                                           &  93.09   & \multicolumn{1}{l|}{94.53}                                             &  94.56    \\
			
			3   &                                                                                                                      
			550644 & \multicolumn{1}{l|}{93.06}                                           & 93.13     & \multicolumn{1}{l|}{94.56}                                             &  94.58  \\
			
			4   &                                                                                                                      
			550852 & \multicolumn{1}{l|}{93.10}                                           &   93.22   & \multicolumn{1}{l|}{94.55}                                             & 94.51  \\
			
			5  &                                                                                                                      
			550907 & \multicolumn{1}{l|}{93.06}                                           
			&   93.14  & \multicolumn{1}{l|}{94.51}                                             &    94.52     \\ \hline                                
		\end{tabular}
	\label{tabel_10}
	\end{table}
	
	%\begin{figure}[h]
	%  \centering
	%  \large
	%  \includegraphics[width=.8\linewidth]{Table1.jpg}
	%  \caption{}
	%  \Description{nlp}
	%\end{figure}

	%From the sentences of Table some wrongly tagged words are found such as the words like "Au" is tagged as CCONJ, whereas it should have been tagged as DET because the word When they are used as prenominal modifiers, they are tagged as determiners and also word "prati" is tagged as PRON but the correct tag that should have been tagged is ADP because the token used after a noun either as a case marker attached with the noun or other such phrases, it is tagged as a adposition.
	
	\section{Conclusion and Future Work}
	\label{conclusion}

In this work, we have presented various methods for developing part of speech tagging for the Odia language that are based on statistical and deep learning-based approaches. Both a statistical method like conditional random field (CRF) and a deep learning-based technique like convolutional neural network (CNN) and bidirectional long short term memory (Bi-LSTM) were considered in this research. Models such as CRF, Bi-LSTM, and CNN are trained using the ILCI corpus with BIS and UD tagsets.  We experimented with different feature set inputs to the CRF model, observed the impact of the constructed feature set, and achieved a 92.08 percent accuracy. We gathered raw corpora from a variety of online resources and Odia-specific websites, and then trained neural word embedding for Odia words. These word vectors are integrated into the deep learning model in order to reduce the number of out-of-vocabulary words. The deep learning-based model is comprised of a Bi-LSTM network, CNN network, CRF layer, character sequence data, and a pre-trained word vector. Bi-LSTM model with pre-trained word embedding and character sequence feature extracted by CNN yielded 94.48 percent accuracy. Moreover, we have also achieved an accuracy of 94.58\% by using the label information of both the models input to Bi-LSTM model. In comparison to other existing studies on the Odia language, the proposed approaches produce more accurate results. 

There are significant possible directions for future research. Since our model does not require knowledge that is task- or domain-specific, one important direction would be to apply it to data from other domains. In addition, computational linguists dealing with low-resource languages suffer with a lack of resources, such as labelled corpora. In the future, we would want to investigate this possibility to generate more labelled datasets across all domains. Another area of research is to e extending this study and applying deep learning to other natural language processing (NLP) problems, such as name entity recognition, chunking, parsing, and machine translation.

	\clearpage
	\bibliographystyle{ACM-Reference-Format}
	\bibliography{Odia_Tagger}


\begin{thebibliography}{35}

%%% the Web address.

\ifx \showCODEN    \undefined \def \showCODEN     #1{\unskip}     \fi
\ifx \showDOI      \undefined \def \showDOI       #1{#1}\fi
\ifx \showISBNx    \undefined \def \showISBNx     #1{\unskip}     \fi
\ifx \showISBNxiii \undefined \def \showISBNxiii  #1{\unskip}     \fi
\ifx \showISSN     \undefined \def \showISSN      #1{\unskip}     \fi
\ifx \showLCCN     \undefined \def \showLCCN      #1{\unskip}     \fi
\ifx \shownote     \undefined \def \shownote      #1{#1}          \fi
\ifx \showarticletitle \undefined \def \showarticletitle #1{#1}   \fi
\ifx \showURL      \undefined \def \showURL       {\relax}        \fi
% invisible to TeX
\providecommand\bibfield[2]{#2}
\providecommand\bibinfo[2]{#2}
\providecommand\natexlab[1]{#1}
\providecommand\showeprint[2][]{arXiv:#2}

\bibitem[Alam et~al\mbox{.}(2016)]%
        {alam2016bidirectional}
\bibfield{author}{\bibinfo{person}{Firoj Alam}, \bibinfo{person}{Shammur~Absar
  Chowdhury}, {and} \bibinfo{person}{Sheak Rashed~Haider Noori}.}
  \bibinfo{year}{2016}\natexlab{}.
\newblock \showarticletitle{Bidirectional lstms—crfs networks for bangla pos
  tagging}. In \bibinfo{booktitle}{\emph{2016 19th International Conference on
  Computer and Information Technology (ICCIT)}}. IEEE,
  \bibinfo{pages}{377--382}.
\newblock


\bibitem[Bojanowski et~al\mbox{.}(2017)]%
        {bojanowski2017enriching}
\bibfield{author}{\bibinfo{person}{Piotr Bojanowski}, \bibinfo{person}{Edouard
  Grave}, \bibinfo{person}{Armand Joulin}, {and} \bibinfo{person}{Tomas
  Mikolov}.} \bibinfo{year}{2017}\natexlab{}.
\newblock \showarticletitle{Enriching word vectors with subword information}.
\newblock \bibinfo{journal}{\emph{Transactions of the association for
  computational linguistics}}  \bibinfo{volume}{5} (\bibinfo{year}{2017}),
  \bibinfo{pages}{135--146}.
\newblock


\bibitem[Brill(1995)]%
        {brill1995transformation}
\bibfield{author}{\bibinfo{person}{Eric Brill}.}
  \bibinfo{year}{1995}\natexlab{}.
\newblock \showarticletitle{Transformation-based error-driven learning and
  natural language processing: A case study in part-of-speech tagging}.
\newblock \bibinfo{journal}{\emph{Computational linguistics}}
  \bibinfo{volume}{21}, \bibinfo{number}{4} (\bibinfo{year}{1995}),
  \bibinfo{pages}{543--565}.
\newblock


\bibitem[Chandra et~al\mbox{.}(2014)]%
        {chandra2014various}
\bibfield{author}{\bibinfo{person}{Nitish Chandra}, \bibinfo{person}{Sudhakar
  Kumawat}, {and} \bibinfo{person}{Vinayak Srivastava}.}
  \bibinfo{year}{2014}\natexlab{}.
\newblock \showarticletitle{Various tagsets for indian languages and their
  performance in part of speech tagging Proceedings of 5 th IRF International
  Conference}.
\newblock \bibinfo{journal}{\emph{Chennai, 23rd March}} (\bibinfo{year}{2014}).
\newblock


\bibitem[Cutting et~al\mbox{.}(1992)]%
        {cutting1992}
\bibfield{author}{\bibinfo{person}{Doug Cutting}, \bibinfo{person}{Julian
  Kupiec}, \bibinfo{person}{Jan Pedersen}, {and} \bibinfo{person}{Penelope
  Sibun}.} \bibinfo{year}{1992}\natexlab{}.
\newblock \showarticletitle{A Practical Part-of-Speech Tagger}. In
  \bibinfo{booktitle}{\emph{Proceedings of the Third Conference on Applied
  Natural Language Processing}} (Trento, Italy) \emph{(\bibinfo{series}{ANLC
  '92})}. \bibinfo{publisher}{Association for Computational Linguistics},
  \bibinfo{address}{USA}, \bibinfo{pages}{133–140}.
\newblock
\urldef\tempurl%
\url{https://doi.org/10.3115/974499.974523}
\showDOI{\tempurl}


\bibitem[Das and Patnaik(2014)]%
        {das2014novel}
\bibfield{author}{\bibinfo{person}{Bishwa~Ranjan Das} {and}
  \bibinfo{person}{Srikanta Patnaik}.} \bibinfo{year}{2014}\natexlab{}.
\newblock \showarticletitle{A novel approach for Odia part of speech tagging
  using artificial neural network}. In \bibinfo{booktitle}{\emph{Proceedings of
  the International Conference on Frontiers of Intelligent Computing: Theory
  and Applications (FICTA) 2013}}. Springer, \bibinfo{pages}{147--154}.
\newblock


\bibitem[Das et~al\mbox{.}(2015)]%
        {das2015part}
\bibfield{author}{\bibinfo{person}{Bishwa~Ranjan Das},
  \bibinfo{person}{Smrutirekha Sahoo}, \bibinfo{person}{Chandra~Sekhar Panda},
  {and} \bibinfo{person}{Srikanta Patnaik}.} \bibinfo{year}{2015}\natexlab{}.
\newblock \showarticletitle{Part of speech tagging in odia using support vector
  machine}.
\newblock \bibinfo{journal}{\emph{Procedia Computer Science}}
  \bibinfo{volume}{48} (\bibinfo{year}{2015}), \bibinfo{pages}{507--512}.
\newblock


\bibitem[De~Marneffe et~al\mbox{.}(2014)]%
        {de2014universal}
\bibfield{author}{\bibinfo{person}{Marie-Catherine De~Marneffe},
  \bibinfo{person}{Timothy Dozat}, \bibinfo{person}{Natalia Silveira},
  \bibinfo{person}{Katri Haverinen}, \bibinfo{person}{Filip Ginter},
  \bibinfo{person}{Joakim Nivre}, {and} \bibinfo{person}{Christopher~D
  Manning}.} \bibinfo{year}{2014}\natexlab{}.
\newblock \showarticletitle{Universal Stanford dependencies: A cross-linguistic
  typology}. In \bibinfo{booktitle}{\emph{Proceedings of the Ninth
  International Conference on Language Resources and Evaluation (LREC'14)}}.
  \bibinfo{pages}{4585--4592}.
\newblock


\bibitem[De~Marneffe and Manning(2008)]%
        {de2008stanford}
\bibfield{author}{\bibinfo{person}{Marie-Catherine De~Marneffe} {and}
  \bibinfo{person}{Christopher~D Manning}.} \bibinfo{year}{2008}\natexlab{}.
\newblock \showarticletitle{The Stanford typed dependencies representation}. In
  \bibinfo{booktitle}{\emph{Coling 2008: proceedings of the workshop on
  cross-framework and cross-domain parser evaluation}}. \bibinfo{pages}{1--8}.
\newblock


\bibitem[Dhanalakshmi et~al\mbox{.}(2009)]%
        {dhanalakshmi2009tamil}
\bibfield{author}{\bibinfo{person}{V Dhanalakshmi}, \bibinfo{person}{G
  Shivapratap}, {and} \bibinfo{person}{Rajendran~S Soman~Kp}.}
  \bibinfo{year}{2009}\natexlab{}.
\newblock \showarticletitle{Tamil POS tagging using linear programming}.
\newblock  (\bibinfo{year}{2009}).
\newblock


\bibitem[Dos~Santos and Zadrozny(2014)]%
        {dos2014learning}
\bibfield{author}{\bibinfo{person}{Cicero Dos~Santos} {and}
  \bibinfo{person}{Bianca Zadrozny}.} \bibinfo{year}{2014}\natexlab{}.
\newblock \showarticletitle{Learning character-level representations for
  part-of-speech tagging}. In \bibinfo{booktitle}{\emph{International
  Conference on Machine Learning}}. PMLR, \bibinfo{pages}{1818--1826}.
\newblock


\bibitem[Ekbal et~al\mbox{.}(2007)]%
        {ekbal2007bengali}
\bibfield{author}{\bibinfo{person}{Asif Ekbal}, \bibinfo{person}{Rejwanul
  Haque}, {and} \bibinfo{person}{Sivaji Bandyopadhyay}.}
  \bibinfo{year}{2007}\natexlab{}.
\newblock \showarticletitle{Bengali part of speech tagging using conditional
  random field}. In \bibinfo{booktitle}{\emph{Proceedings of seventh
  international symposium on natural language processing (SNLP2007)}}.
  \bibinfo{pages}{131--136}.
\newblock


\bibitem[Harris(1962)]%
        {harris1962string}
\bibfield{author}{\bibinfo{person}{Zellig Harris}.}
  \bibinfo{year}{1962}\natexlab{}.
\newblock \showarticletitle{String analysis of language structure}.
\newblock \bibinfo{journal}{\emph{Mouton and Co., The Hague}}
  (\bibinfo{year}{1962}).
\newblock


\bibitem[Huang et~al\mbox{.}(2015)]%
        {huang2015bidirectional}
\bibfield{author}{\bibinfo{person}{Zhiheng Huang}, \bibinfo{person}{Wei Xu},
  {and} \bibinfo{person}{Kai Yu}.} \bibinfo{year}{2015}\natexlab{}.
\newblock \showarticletitle{Bidirectional LSTM-CRF models for sequence
  tagging}.
\newblock \bibinfo{journal}{\emph{arXiv preprint arXiv:1508.01991}}
  (\bibinfo{year}{2015}).
\newblock


\bibitem[Krishnan et~al\mbox{.}(2017)]%
        {krishnan2017character}
\bibfield{author}{\bibinfo{person}{KS~Gokul Krishnan}, \bibinfo{person}{A
  Pooja}, \bibinfo{person}{M~Anand Kumar}, {and} \bibinfo{person}{KP Soman}.}
  \bibinfo{year}{2017}\natexlab{}.
\newblock \showarticletitle{Character based bidirectional LSTM for
  disambiguating tamil part-of-speech categories}.
\newblock \bibinfo{journal}{\emph{Int. J. Control Theory Appl}}
  \bibinfo{volume}{229} (\bibinfo{year}{2017}), \bibinfo{pages}{235}.
\newblock


\bibitem[Kunchukuttan et~al\mbox{.}(2020)]%
        {kunchukuttan2020indicnlpcorpus}
\bibfield{author}{\bibinfo{person}{Anoop Kunchukuttan},
  \bibinfo{person}{Divyanshu Kakwani}, \bibinfo{person}{Satish Golla},
  \bibinfo{person}{Gokul N.C.}, \bibinfo{person}{Avik Bhattacharyya},
  \bibinfo{person}{Mitesh~M. Khapra}, {and} \bibinfo{person}{Pratyush Kumar}.}
  \bibinfo{year}{2020}\natexlab{}.
\newblock \showarticletitle{AI4Bharat-IndicNLP Corpus: Monolingual Corpora and
  Word Embeddings for Indic Languages}.
\newblock \bibinfo{journal}{\emph{arXiv preprint arXiv:2005.00085}}
  (\bibinfo{year}{2020}).
\newblock


\bibitem[Lafferty et~al\mbox{.}(2001)]%
        {lafferty2001conditional}
\bibfield{author}{\bibinfo{person}{John Lafferty}, \bibinfo{person}{Andrew
  McCallum}, {and} \bibinfo{person}{Fernando~CN Pereira}.}
  \bibinfo{year}{2001}\natexlab{}.
\newblock \showarticletitle{Conditional random fields: Probabilistic models for
  segmenting and labeling sequence data}.
\newblock  (\bibinfo{year}{2001}).
\newblock


\bibitem[Ling et~al\mbox{.}(2015)]%
        {ling2015finding}
\bibfield{author}{\bibinfo{person}{Wang Ling}, \bibinfo{person}{Tiago
  Lu{\'\i}s}, \bibinfo{person}{Lu{\'\i}s Marujo},
  \bibinfo{person}{Ram{\'o}n~Fernandez Astudillo}, \bibinfo{person}{Silvio
  Amir}, \bibinfo{person}{Chris Dyer}, \bibinfo{person}{Alan~W Black}, {and}
  \bibinfo{person}{Isabel Trancoso}.} \bibinfo{year}{2015}\natexlab{}.
\newblock \showarticletitle{Finding function in form: Compositional character
  models for open vocabulary word representation}.
\newblock \bibinfo{journal}{\emph{arXiv preprint arXiv:1508.02096}}
  (\bibinfo{year}{2015}).
\newblock


\bibitem[Mitkov(2022)]%
        {mitkov2022oxford}
\bibfield{author}{\bibinfo{person}{Ruslan Mitkov}.}
  \bibinfo{year}{2022}\natexlab{}.
\newblock \bibinfo{booktitle}{\emph{The Oxford handbook of computational
  linguistics}}.
\newblock \bibinfo{publisher}{Oxford University Press}.
\newblock


\bibitem[Nivre et~al\mbox{.}(2016)]%
        {nivre2016universal}
\bibfield{author}{\bibinfo{person}{Joakim Nivre},
  \bibinfo{person}{Marie-Catherine De~Marneffe}, \bibinfo{person}{Filip
  Ginter}, \bibinfo{person}{Yoav Goldberg}, \bibinfo{person}{Jan Hajic},
  \bibinfo{person}{Christopher~D Manning}, \bibinfo{person}{Ryan McDonald},
  \bibinfo{person}{Slav Petrov}, \bibinfo{person}{Sampo Pyysalo},
  \bibinfo{person}{Natalia Silveira}, {et~al\mbox{.}}}
  \bibinfo{year}{2016}\natexlab{}.
\newblock \showarticletitle{Universal dependencies v1: A multilingual treebank
  collection}. In \bibinfo{booktitle}{\emph{Proceedings of the Tenth
  International Conference on Language Resources and Evaluation (LREC'16)}}.
  \bibinfo{pages}{1659--1666}.
\newblock


\bibitem[Nivre et~al\mbox{.}(2020)]%
        {nivre2020universal}
\bibfield{author}{\bibinfo{person}{Joakim Nivre},
  \bibinfo{person}{Marie-Catherine de Marneffe}, \bibinfo{person}{Filip
  Ginter}, \bibinfo{person}{Jan Haji{\v{c}}}, \bibinfo{person}{Christopher~D
  Manning}, \bibinfo{person}{Sampo Pyysalo}, \bibinfo{person}{Sebastian
  Schuster}, \bibinfo{person}{Francis Tyers}, {and} \bibinfo{person}{Daniel
  Zeman}.} \bibinfo{year}{2020}\natexlab{}.
\newblock \showarticletitle{Universal Dependencies v2: An evergrowing
  multilingual treebank collection}.
\newblock \bibinfo{journal}{\emph{arXiv preprint arXiv:2004.10643}}
  (\bibinfo{year}{2020}).
\newblock


\bibitem[Ojha et~al\mbox{.}(2015)]%
        {ojha2015training}
\bibfield{author}{\bibinfo{person}{Atul~Ku Ojha}, \bibinfo{person}{Pitambar
  Behera}, \bibinfo{person}{Srishti Singh}, {and} \bibinfo{person}{Girish~N
  Jha}.} \bibinfo{year}{2015}\natexlab{}.
\newblock \showarticletitle{Training \& evaluation of POS taggers in Indo-Aryan
  languages: a case of Hindi, Odia and Bhojpuri}. In
  \bibinfo{booktitle}{\emph{the proceedings of 7th language \& technology
  conference: human language technologies as a challenge for computer science
  and linguistics}}. \bibinfo{pages}{524--529}.
\newblock


\bibitem[Parida et~al\mbox{.}(2020a)]%
        {parida2020odiencorp}
\bibfield{author}{\bibinfo{person}{Shantipriya Parida},
  \bibinfo{person}{Ond{\v{r}}ej Bojar}, {and} \bibinfo{person}{Satya~Ranjan
  Dash}.} \bibinfo{year}{2020}\natexlab{a}.
\newblock \showarticletitle{OdiEnCorp: Odia--English and Odia-Only Corpus for
  Machine Translation}.
\newblock In \bibinfo{booktitle}{\emph{Smart Intelligent Computing and
  Applications}}. \bibinfo{publisher}{Springer}, \bibinfo{pages}{495--504}.
\newblock


\bibitem[Parida et~al\mbox{.}(2020b)]%
        {parida2020odiencorp2}
\bibfield{author}{\bibinfo{person}{Shantipriya Parida},
  \bibinfo{person}{Satya~Ranjan Dash}, \bibinfo{person}{Ond{\v{r}}ej Bojar},
  \bibinfo{person}{Petr Motlicek}, \bibinfo{person}{Priyanka Pattnaik}, {and}
  \bibinfo{person}{Debasish~Kumar Mallick}.} \bibinfo{year}{2020}\natexlab{b}.
\newblock \showarticletitle{OdiEnCorp 2.0: Odia-English Parallel Corpus for
  Machine Translation}. In \bibinfo{booktitle}{\emph{Proceedings of the
  WILDRE5--5th Workshop on Indian Language Data: Resources and Evaluation}}.
  \bibinfo{pages}{14--19}.
\newblock


\bibitem[Pattnaik et~al\mbox{.}(2020)]%
        {pattnaik2020semi}
\bibfield{author}{\bibinfo{person}{Sagarika Pattnaik},
  \bibinfo{person}{Ajit~Kumar Nayak}, {and} \bibinfo{person}{Srikanta
  Patnaik}.} \bibinfo{year}{2020}\natexlab{}.
\newblock \showarticletitle{A Semi-supervised Learning of HMM to Build a POS
  Tagger for a Low Resourced Language}.
\newblock \bibinfo{journal}{\emph{Journal of information and communication
  convergence engineering}} \bibinfo{volume}{18}, \bibinfo{number}{4}
  (\bibinfo{year}{2020}), \bibinfo{pages}{207--215}.
\newblock


\bibitem[Petrov et~al\mbox{.}(2011)]%
        {petrov2011universal}
\bibfield{author}{\bibinfo{person}{Slav Petrov}, \bibinfo{person}{Dipanjan
  Das}, {and} \bibinfo{person}{Ryan McDonald}.}
  \bibinfo{year}{2011}\natexlab{}.
\newblock \showarticletitle{A universal part-of-speech tagset}.
\newblock \bibinfo{journal}{\emph{arXiv preprint arXiv:1104.2086}}
  (\bibinfo{year}{2011}).
\newblock


\bibitem[Plank et~al\mbox{.}(2016)]%
        {plank2016multilingual}
\bibfield{author}{\bibinfo{person}{Barbara Plank}, \bibinfo{person}{Anders
  S{\o}gaard}, {and} \bibinfo{person}{Yoav Goldberg}.}
  \bibinfo{year}{2016}\natexlab{}.
\newblock \showarticletitle{Multilingual part-of-speech tagging with
  bidirectional long short-term memory models and auxiliary loss}.
\newblock \bibinfo{journal}{\emph{arXiv preprint arXiv:1604.05529}}
  (\bibinfo{year}{2016}).
\newblock


\bibitem[Priyadarshi and Saha(2020)]%
        {priyadarshi2020towards}
\bibfield{author}{\bibinfo{person}{Ankur Priyadarshi} {and}
  \bibinfo{person}{Sujan~Kumar Saha}.} \bibinfo{year}{2020}\natexlab{}.
\newblock \showarticletitle{Towards the first Maithili part of speech tagger:
  Resource creation and system development}.
\newblock \bibinfo{journal}{\emph{Computer Speech \& Language}}
  \bibinfo{volume}{62} (\bibinfo{year}{2020}), \bibinfo{pages}{101054}.
\newblock


\bibitem[Ramesh et~al\mbox{.}(2022)]%
        {ramesh2022samanantar}
\bibfield{author}{\bibinfo{person}{Gowtham Ramesh}, \bibinfo{person}{Sumanth
  Doddapaneni}, \bibinfo{person}{Aravinth Bheemaraj}, \bibinfo{person}{Mayank
  Jobanputra}, \bibinfo{person}{Raghavan AK}, \bibinfo{person}{Ajitesh Sharma},
  \bibinfo{person}{Sujit Sahoo}, \bibinfo{person}{Harshita Diddee},
  \bibinfo{person}{Divyanshu Kakwani}, \bibinfo{person}{Navneet Kumar},
  {et~al\mbox{.}}} \bibinfo{year}{2022}\natexlab{}.
\newblock \showarticletitle{Samanantar: The largest publicly available parallel
  corpora collection for 11 indic languages}.
\newblock \bibinfo{journal}{\emph{Transactions of the Association for
  Computational Linguistics}}  \bibinfo{volume}{10} (\bibinfo{year}{2022}),
  \bibinfo{pages}{145--162}.
\newblock


\bibitem[Schmid(1994)]%
        {schmid1994part}
\bibfield{author}{\bibinfo{person}{Helmut Schmid}.}
  \bibinfo{year}{1994}\natexlab{}.
\newblock \showarticletitle{Part-of-speech tagging with neural networks}.
\newblock \bibinfo{journal}{\emph{arXiv preprint cmp-lg/9410018}}
  (\bibinfo{year}{1994}).
\newblock


\bibitem[Shrivastava and Bhattacharyya(2008)]%
        {shrivastava2008hindi}
\bibfield{author}{\bibinfo{person}{Manish Shrivastava} {and}
  \bibinfo{person}{Pushpak Bhattacharyya}.} \bibinfo{year}{2008}\natexlab{}.
\newblock \showarticletitle{Hindi POS tagger using naive stemming: harnessing
  morphological information without extensive linguistic knowledge}. In
  \bibinfo{booktitle}{\emph{International Conference on NLP (ICON08), Pune,
  India}}.
\newblock


\bibitem[Suraksha et~al\mbox{.}(2017)]%
        {suraksha2017part}
\bibfield{author}{\bibinfo{person}{NM Suraksha}, \bibinfo{person}{K Reshma},
  {and} \bibinfo{person}{KM~Shiva Kumar}.} \bibinfo{year}{2017}\natexlab{}.
\newblock \showarticletitle{Part-of-speech tagging and parsing of Kannada text
  using Conditional Random Fields (CRFs)}. In \bibinfo{booktitle}{\emph{2017
  International Conference on Intelligent Computing and Control (I2C2)}}. IEEE,
  \bibinfo{pages}{1--5}.
\newblock


\bibitem[Warjri et~al\mbox{.}(2019)]%
        {warjri2019identification}
\bibfield{author}{\bibinfo{person}{Sunita Warjri}, \bibinfo{person}{Partha
  Pakray}, \bibinfo{person}{Saralin Lyngdoh}, {and} \bibinfo{person}{Arnab
  Kumar~Maji}.} \bibinfo{year}{2019}\natexlab{}.
\newblock \showarticletitle{Identification of pos tag for khasi language based
  on hidden markov model pos tagger}.
\newblock \bibinfo{journal}{\emph{Computaci{\'o}n y Sistemas}}
  \bibinfo{volume}{23}, \bibinfo{number}{3} (\bibinfo{year}{2019}),
  \bibinfo{pages}{795--802}.
\newblock


\bibitem[Warjri et~al\mbox{.}(2021)]%
        {warjri2021part}
\bibfield{author}{\bibinfo{person}{Sunita Warjri}, \bibinfo{person}{Partha
  Pakray}, \bibinfo{person}{Saralin~A Lyngdoh}, {and}
  \bibinfo{person}{Arnab~Kumar Maji}.} \bibinfo{year}{2021}\natexlab{}.
\newblock \showarticletitle{Part-of-Speech (POS) Tagging Using Deep
  Learning-Based Approaches on the Designed Khasi POS Corpus}.
\newblock \bibinfo{journal}{\emph{Transactions on Asian and Low-Resource
  Language Information Processing}} \bibinfo{volume}{21}, \bibinfo{number}{3}
  (\bibinfo{year}{2021}), \bibinfo{pages}{1--24}.
\newblock


\bibitem[Zeman and Resnik(2008)]%
        {zeman2008cross}
\bibfield{author}{\bibinfo{person}{Daniel Zeman} {and} \bibinfo{person}{Philip
  Resnik}.} \bibinfo{year}{2008}\natexlab{}.
\newblock \showarticletitle{Cross-language parser adaptation between related
  languages}. In \bibinfo{booktitle}{\emph{Proceedings of the IJCNLP-08
  Workshop on NLP for Less Privileged Languages}}.
\newblock


\end{thebibliography}
	
\end{document}